\pdfoutput=1

\documentclass{article} % For LaTeX2e

\usepackage[dvipsnames]{xcolor}
\usepackage{nips12submit_e,times}

\usepackage{amsmath}
\usepackage{amsthm}
\usepackage{amssymb}
\usepackage{amsfonts}
\usepackage{graphicx}
\usepackage{verbatim}
\usepackage{bm}
\usepackage{mathrsfs}
\usepackage{enumerate}
\usepackage{wrapfig}

\title{Minimax Multi-Task Learning and a Generalized Loss-Compositional Paradigm for MTL}

\author{
Nishant A.~Mehta$^\dag$, Dongryeol Lee\thanks{Work completed while at Georgia Institute of Technology}, Alexander G.~Gray$^\dag$ \\
\texttt{niche@cc.gatech.edu, drselee@gmail.com, agray@cc.gatech.edu} \\
$^\dag$ College of Computing, Georgia Insitute of Technology,  
Atlanta, GA 30332, USA \\
$^*$ GE Global Research, 
Niskayuna, NY 12309, USA 
}

\nipsfinalcopy % Uncomment for camera-ready version

\newcommand{\E}{\mathsf{E}}

\newcommand{\real}{\mathbb{R}}

\newcommand{\hypspace}{\mathcal{H}}
\newcommand{\X}{\mathcal{X}}
\newcommand{\Y}{\mathcal{Y}}

\newcommand{\rademacher}{\mathcal{R}}

\newtheorem{theorem}{Theorem}

\newtheorem{lemma}{Lemma}

\begin{document}

\maketitle

\begin{abstract}
Since its inception, the modus operandi of multi-task learning (MTL) has been to minimize the task-wise mean of the empirical risks.  
We introduce a generalized loss-compositional paradigm for MTL that includes a spectrum of formulations as a subfamily. One endpoint of this spectrum is minimax MTL: a new MTL formulation that minimizes the maximum of the tasks' empirical risks. Via a certain relaxation of minimax MTL, we obtain a continuum of MTL formulations spanning minimax MTL and classical MTL. The full paradigm itself is loss-compositional, operating on the vector of empirical risks. It incorporates minimax MTL, its relaxations, and many new MTL formulations as special cases. We show theoretically that minimax MTL tends to avoid worst case outcomes on newly drawn test tasks in the learning to learn (LTL) test setting. The results of several MTL formulations on synthetic and real problems in the MTL and LTL test settings are encouraging. 
\end{abstract}

\section{Introduction}

The essence of machine learning is to exploit what we observe in order to form accurate predictors of what we cannot. 
A multi-task learning (MTL) algorithm learns an inductive bias to learn several tasks together. 
MTL is incredibly pervasive in machine learning: it has natural connections to random effects models \cite{yu2009large}; user preference prediction (including collaborative filtering) can be framed as MTL \cite{zhang2011generalizing}; multi-class classification admits the popular \emph{one-vs-all} and \emph{all-pairs} MTL reductions; and MTL admits provably good learning in settings where single-task learning is hopeless \cite{baxter2000model,maurer2009transfer}. 
But if we see examples from a random set of tasks today, which of these tasks will matter tomorrow? 
Not knowing in the present what challenges nature has in store for the future, a sensible strategy is to mitigate the worst case by ensuring some minimum proficiency on each task. 

Consider a simple learning scenario: 
A music preference prediction company is in the business of predicting what 5-star ratings different users would assign to songs. 
At training time, the company learns a shared representation for predicting the users' song ratings by pooling together the company's limited data on each user's preferences. Given this learned representation, a separate predictor for each user can be trained very quickly. At test time, the environment draws a user according to some (possibly randomized) rule and solicits from the company a prediction of that user's preference for a particular song. The environment may also ask for predictions about new users, described by a few ratings each, and so the company must leverage its existing representation to rapidly learn new predictors and produce ratings for these new users. 

Classically, multi-task learning has sought to minimize the (regularized) sum of the empirical risks over a set of tasks. 
In this way, classical MTL implicitly assumes that once the learner has been trained, it will be tested on test tasks drawn uniformly at random from the empirical task distribution of the training tasks. 
Notably, there are several reasons why classical MTL may not be ideal:
\begin{list}{\labelitemi}{\leftmargin=1.5em}
\item While at training time the usual flavor of MTL commits to a fixed distribution over users (typically either uniform or proportional to the number of ratings available for each user), at test time there is no guarantee what user distribution we will encounter. In fact, there may not exist any fixed user distribution: the sequence of users for which ratings are elicited could be adversarial. 
\item Even in the case when the distribution over tasks is not adversarial, it may be in the interest of the music preference prediction company to guarantee some minimum level of accuracy per user in order to minimize negative feedback and a potential loss of business, rather than maximing the mean level of accuracy over all users. 
\item Whereas minimizing the average prediction error is very much a teleological endeavor, typically at the expense of some locally egregious outcomes, minimizing the worst-case prediction error respects a notion of fairness to all tasks (or people). 
\end{list}

This work introduces \emph{minimax multi-task learning} as a response to the above scenario.\footnote{Note that minimax MTL does not refer to the \emph{minimax estimators} of statistical decision theory.}
In addition, we cast a spectrum of multi-task learning. At one end of the spectrum lies minimax MTL, and departing from this point progressively relaxes the ``hardness'' of the maximum until full relaxation reaches the second endpoint and recovers classical MTL. 
We further sculpt a generalized loss-compositional paradigm for MTL which includes this spectrum and several other new MTL formulations. This paradigm equally applies to the problem of \emph{learning to learn} (LTL), in which the goal is to learn a hypothesis space from a set of training tasks such that this representation admits good hypotheses on future tasks. In truth, MTL and LTL typically are handled equivalently at training time --- this work will be no exception --- and they diverge only in their test settings and hence the learning theoretic inquiries they inspire.

\paragraph{Contributions.}
The first contribution of this work is to introduce minimax MTL and a continuum of relaxations. 
Second, we introduce a generalized loss-compositional paradigm for MTL which admits a number of new MTL formulations and also includes classical MTL as a special case. 
Third, we empirically evaluate the performance of several MTL formulations from this paradigm in the multi-task learning and learning to learn settings, under the task-wise maximum test risk and task-wise mean test risk criteria, on four datasets (one synthetic, three real). 
Finally, Theorem \ref{thm:max-true-risk} is the core theoretical contribution of this work and shows the following: If it is possible to obtain maximum empirical risk across a set of training tasks below some level $\gamma$, then it is likely that the maximum true risk obtained by the learner on a new task is bounded by roughly $\gamma$. 
Hence, if the goal is to minimize the worst case outcome over new tasks, the theory suggests minimizing the maximum of the empirical risks across the training tasks rather than their mean.  

In the next section, we recall the settings of multi-task learning and learning to learn, formally introduce minimax MTL, and motivate it theoretically. 
In Section \ref{sec:paradigm}, we introduce a continuously parametrized family of minimax MTL relaxations and the new generalized loss-compositional paradigm. 
Section \ref{sec:experiments} presents an empirical evaluation of various MTL/LTL formulations with different models on four datasets. Finally, we close with a discussion.

\section{Minimax multi-task learning}
We begin with a promenade through the basic MTL and LTL setups, with an effort to abide by the notation introduced by Baxter \cite{baxter2000model}. Throughout the rest of the paper, each labeled example $(x,y)$ will live in $\X \times \Y$ for input instance $x$ and label $y$. Typical choices of $\X$ include $\real^n$ or a compact subset thereof, while $\Y$ typically is a compact subset of $\real$ or the binary $\{-1,1\}$. In addition, define a loss function $\ell: \real \times \Y \rightarrow \real_+$. For simplicity, this work considers $\ell_2$ loss (squared loss) $\ell(y',y) = (y' - y)^2$ for regression and hinge loss  $\ell(y', y) = \max\{0, 1 - y' y\}$ for classification. 

MTL and LTL often are framed as applying an inductive bias to learn a common hypothesis space, selected from a fixed family of hypothesis spaces, and thereafter learning from this hypothesis space a hypothesis for each task observed at training time. 
It will be useful to formalize the various sets and elements present in the preceding statement. Let $\mathbb{H}$ be a family of hypothesis spaces. Any hypothesis space $\hypspace \in \mathbb{H}$ itself is a set of hypotheses; each hypothesis $h \in \hypspace$ is a map $h: \X \rightarrow \real$.

\paragraph{Learning to learn.} 
In learning to learn, the goal is to achieve inductive transfer to learn the best $\hypspace$ from $\mathbb{H}$. Unlike in MTL, there is a notion of an \emph{environment} of tasks: an unknown probability measure $Q$ over a space of task probability measures $\mathcal{P}$. The goal is to find the optimal representation via the objective
\begin{align}
\inf_{\hypspace \in \mathbb{H}} \E_{P \sim Q} \inf_{h \in \hypspace} \E_{(x,y) \sim P} \ell(y, h(x)) .
\label{eqn:ltl-goal}
\end{align}
In practice, $T$ (unobservable) training task probability measures $P_1, \ldots, P_T \in \mathcal{P}$ are drawn iid from $Q$, and from each task $t$ a set of $m$ examples are drawn iid from $P_t$.

\paragraph{Multi-task learning.} 
Whereas in learning to learn there is a distribution over tasks, in multi-task learning there is a fixed, finite set of tasks indexed by $[T] := \{1, \ldots, T\}$. Each task $t \in [T]$ is coupled with a fixed but unknown probability measure $P_t$. 
Classically, the goal of MTL is to minimize the expected loss at test time under the uniform distribution on $[T]$:
\begin{align}
\inf_{\hypspace \in \mathbb{H}} \,\, \frac{1}{T} \sum_{t \in [T]} \inf_{h \in \hypspace}  \E_{(x,y) \sim P_t} \ell(y, h(x)) .
\label{eqn:mtl-goal}
\end{align}
Notably, this objective is equivalent to \eqref{eqn:ltl-goal} when $Q$ is the uniform distribution on $\{P_1, \ldots, P_T\}$. In terms of the data generation model,  MTL differs from LTL since the tasks are fixed; however, just as in LTL, from each task $t$ a set of $m$ examples are drawn iid from $P_t$ .

\subsection{Minimax MTL}
A natural generalization of classical MTL results by introducing a prior distribution $\pi$ over the index set of tasks $[T]$. Given $\pi$, the (idealized) objective of this generalized MTL is
\begin{align}
\inf_{\hypspace \in \mathbb{H}} \E_{t \sim \pi} \inf_{h \in \hypspace} \E_{(x,y) \sim P_t} \ell(y, h(x)),
\label{eqn:pi-true-risk}
\end{align}
given only the training data $\{(x_{t,1}, y_{t,1}), \ldots, (x_{t,m}, y_{t,m})\}_{t \in [T]}$. 
The classical MTL objective \eqref{eqn:mtl-goal} equals \eqref{eqn:pi-true-risk} when $\pi$ is taken to be the uniform prior over $[T]$. 
We argue that in many instances, that which is most relevant to minimize is not the expected error under a uniform distribution over tasks, or even any pre-specified $\pi$, but rather the expected error for the worst $\pi$. We propose to minimize the maximum error over tasks under an adversarial choice of $\pi$, yielding the objective:
\begin{align*}
\inf_{\hypspace \in \mathbb{H}} \sup_\pi \E_{t \sim \pi} \inf_{h \in \hypspace} \E_{(x,y) \sim P_t} \ell(y, h(x)) ,
\end{align*}
where the supremum is taken over the $T$-dimensional simplex. 
As the supremum (assuming it is attained) is attained at an extreme point of the simplex, this objective is equivalent to
\begin{align}
\inf_{\hypspace \in \mathbb{H}} \max_{t \in [T]} \inf_{h \in \hypspace} \E_{(x,y) \sim P_t} \ell(y, h(x)) . \label{eqn:mtl-true-minimax}
\end{align}

In practice, we approximate the true objective by using the (regularized) empirical objective:
\begin{align*}
\inf_{\hypspace \in \mathbb{H}} \max_{t \in [T]} \inf_{h \in \hypspace} \sum_{i=1}^m \ell(y_{t,i}, h(x_{t,i})) .
\end{align*}

In the next section, we motivate minimax MTL theoretically by showing that the worst-case performance on future tasks likely will not be much higher than the maximum of the empirical risks for the training tasks. In this short paper, we restrict attention to the case of finite $\mathbb{H}$.

\subsection{A learning to learn bound for  the maximum risk}
\label{sec:theory}

In this subsection, we use the following notation. Let $P^{(1)}, \ldots, P^{(T)}$ be probability measures drawn iid from $Q$, and for $t \in [T]$ let $\mathbf{z}^{(t)}$ be an $m$-sample (a sample of $m$ points) from $P^{(t)}$ with corresponding empirical measure $P_m^{(t)}$. Also, if $P$ is as a probability measure then $P \ell \circ h := \E \ell(y, h(x))$; 
similarly, if $P_m$ is an empirical measure, 
then $P_m \ell \circ h := \frac{1}{m} \sum_{i=1}^m \ell(y_i, h(x_i))$. 

Our focus is the learning to learn setting with a minimax lens: when one learns a representation $\hypspace \in\mathbb{H}$ from multiple training tasks and observes maximum empirical risk $\gamma$, we would like to guarantee that $\hypspace$'s true risk on a newly drawn test task will be bounded by roughly $\gamma$. Such a goal is in striking contrast to the classical emphasis of learning to learn, where the goal is to obtain bounds on $\hypspace$'s expected true risk. Using $\hypspace$'s expected true risk and Markov's inequality, Baxter \cite[the display prior to (25) ]{baxter2000model} showed that the probability that $\hypspace$'s true risk on a newly drawn test task is above some level $\gamma$ decays as the expected true risk over $\gamma$:
\begin{align}
\mathsf{Pr} \left\{ \inf_{h \in \hypspace} P \ell \circ h \geq \gamma \right\} 
\leq \frac{\frac{1}{T} \sum_{t \in [T]} P^{(t)}_m \ell \circ h_t + \varepsilon}{\gamma}
\label{eqn:ltl-markov}
\end{align}
where the size of $\varepsilon$ is controlled by $T$, $m$, and the complexities of certain spaces.

The expected true risk is not of primary interest for controlling the tail of the (random) true risk, and a more direct approach yields a much better bound. In this short paper we restrict the space of representations $\mathbb{H}$ to be finite with cardinality $\mathcal{C}$; in this case, the analysis is particularly simple and illuminates the idea for proving the general case. The next theorem is the main result of this section:
\begin{theorem}
\label{thm:max-true-risk}
Let $|\mathbb{H}| = \mathcal{C}$, and let the loss $\ell$ be $L$-Lipschitz in its second argument and bounded by $B$.  Suppose $T$ tasks $P^{(1)}, \ldots, P^{(T)}$ are drawn iid from $Q$ and from each task $P^{(t)}$ an iid $m$-sample $\mathbf{z}^{(t)}$ is drawn. Suppose there exists $\hypspace \in \mathbb{H}$ such that all $t \in [T]$ satisfy $\min_{h \in \hypspace} P_m^{(t)} \ell \circ h \leq \gamma$. 
Let $P$ be newly drawn probability measure from $Q$. 
Let $\hat{h}$ be the empirical risk minimizer over the test $m$-sample. 
With probability at least $1 - \delta$ with respect to the random draw of the $T$ tasks and their $T$ corresponding $m$-samples:
\begin{align}
\mathsf{Pr} \left\{  P \ell \circ \hat{h} 
> \gamma  + \frac{1}{T} 
   + 2 L \max_{\hypspace \in \mathbb{H}} \rademacher_m(\hypspace) 
   + \sqrt{\frac{8 \log \frac{4}{\delta}}{m}} 
\right\}
\leq \frac{\log \frac{2 \mathcal{C}}{\delta} + \log \lceil B \rceil + \log(T+1)}{T} .
\label{eqn:ltl-direct}
\end{align}
\end{theorem}
In the above, $\rademacher_m(\hypspace)$ is the Rademacher complexity of $\hypspace$ (cf. \cite{bartlett2002rademacher}). 
Critically, in \eqref{eqn:ltl-direct} the probability of observing a task with high true risk decays with $T$, whereas in \eqref{eqn:ltl-markov} the decay is independent of $T$. Hence, when the goal is to minimize the probability of bad performance on future tasks uniformly, this theorem motivates minimizing the \emph{maximum} of the empirical risks as opposed to their mean.

For the proof of Theorem \ref{thm:max-true-risk}, first consider the singleton case $\mathbb{H} = \{\hypspace_1\}$. 
Suppose that for $\gamma$ fixed a priori, the maximum of the empirical risks is bounded by $\gamma$, i.e. 
$
\max_{t \in [T]} \min_{h \in \hypspace_1} P_m^{(t)} \ell \circ h \leq \gamma 
$. 

Let a new probability measure $P$ drawn from $Q$ correspond to a new test task. Suppose the probability of the event $[\min_{h \in \hypspace_1} P_m \ell \circ h > \gamma]$ is at least $\varepsilon$. Then the probability that $\gamma$ bounds all $T$ empirical risks is at most $(1 - \varepsilon)^{T} \leq e^{-T \varepsilon}$. Hence, with probability at least $1 - e^{-T \varepsilon}$:
\begin{align}
\textstyle
\mathsf{Pr} \left\{  \min_{h \in \hypspace_1} P_m \ell \circ h > \gamma  \right\}
\leq \varepsilon .
\end{align}

A simple application of the union bound extends this result for finite $\mathbb{H}$:
\begin{lemma}
\label{lemma:max-empirical-risk}
Under the same conditions as Theorem \ref{thm:max-true-risk}, with probability at least $1 - \delta / 2$ with respect to the random draw of the $T$ tasks and their $T$ corresponding $m$-samples:
{
\begin{align}
\mathsf{Pr} \left\{  \min_{h \in \hypspace} P_m \ell \circ h > \gamma  \right\}
\leq \frac{\log \frac{2 \mathcal{C}}{\delta}}{T} .
\end{align}
}
\end{lemma}
The bound in the lemma states a $1/T$ rate of decay for the probability that the empirical risk obtained by $\hypspace$ on a new task exceeds $\gamma$. 
Next, we relate this empirical risk with the true risk obtained by the empirical risk minimizer. 
Note that at test time $\hypspace$ is fixed and hence independent of any test $m$-sample. 
Then, from by now standard learning theory results of Bartlett and Mendelson \cite{bartlett2002rademacher}:
\begin{lemma}
Take loss $\ell$ as in Theorem \ref{thm:max-true-risk}. With probability at least $1 - \delta / 2$, for all $h \in \hypspace$ uniformly:
\begin{align}
P \ell \circ h \leq P_m \ell \circ h 
+ 2 L \rademacher_m(\hypspace) 
+ \sqrt{(8 \log(4/\delta)) / m} .
\label{eqn:basic-slt}
\end{align}
\label{lemma:basic-slt}
\end{lemma}

In particular, with high probability the true risk of the empirical risk minimizer is not much larger than its empirical risk. 
Theorem \ref{thm:max-true-risk} now follows from Lemmas \ref{lemma:max-empirical-risk} and \ref{lemma:basic-slt} and a union bound over $\gamma \in \Gamma := \{ 0, 1/T, 2/T, \ldots, \lceil B \rceil \}$; 
note that mapping the observed maximum empirical risk $\gamma$ to $\min \{\gamma' \in \Gamma \mid \gamma \leq \gamma'\}$ picks up 
the additional $\frac{1}{T}$ term in \eqref{eqn:ltl-direct}. 

In the next section, we introduce a loss-compositional paradigm for multi-task learning which includes as special cases minimax MTL and classical MTL. 

\section{A generalized loss-compositional paradigm for MTL}
\label{sec:paradigm}
The paradigm can benefit from a bit of notation. 
Given a set of $T$ tasks, we represent the empirical risk for hypothesis $h_t \in \hypspace$ ($ \in \mathbb{H}$) on task $t \in [T]$ as $\hat{\ell}_t(h_t) := \sum_{i=1}^m \ell(y_{t,i}, h_t(x_{t,i}))$. Additionally define a set of hypotheses for multiple tasks $\mathbf{h} := (h_1, \ldots, h_T) \in \hypspace^T$ and the vector of empirical risks $\hat{\bm{\ell}}(\mathbf{h}) := (\hat{\ell}_1(h_1), \ldots, \hat{\ell}_T(h_T))$. 

With this notation set, the proposed loss-compositional paradigm encompasses any regularized minimization of a (typically convex) function $\phi: \real_+^T \rightarrow \real_+$ of the empirical risks:
\begin{align}
\inf_{\hypspace \in \mathbb{H}} \inf_{\mathbf{h} \in \hypspace^T} \phi \bigl(\hat{\bm{\ell}}(\mathbf{h}) \bigr) + \Omega \bigl( (\hypspace, \mathbf{h}) \bigr),
\label{eqn:paradigm}
\end{align}
where $\Omega(\cdot): \mathbb{H} \times \cup_{\hypspace \in \mathbb{H}} \hypspace^T \rightarrow \real_+$ is a regularizer. 

\paragraph{$\bm{\ell}_{\mathbf{p}}$ MTL.}
One notable specialization that is still quite general is the case when $\phi$ is an $\ell_p$-norm, yielding \emph{$\ell_p$ MTL}. This subfamily encompasses classical MTL and many new MTL formulations:
\begin{list}{\labelitemi}{\leftmargin=2em}
\item 
Classical MTL as \emph{$\ell_1$ MTL}: 

\qquad $\displaystyle
\quad \inf_{\hypspace \in \mathbb{H}} \inf_{\mathbf{h} \in \hypspace^T} \frac{1}{T} \sum_{t \in [T]} \hat{\ell}(h_t) + \Omega \bigl( (\hypspace, \mathbf{h}) \bigr) 
\qquad \,\,\,\, \equiv \,\,
\inf_{\hypspace \in \mathbb{H}} \inf_{\mathbf{h} \in \hypspace^T} \frac{1}{T} \| \hat{\bm{\ell}}(\mathbf{h}) \|_1 + \Omega \bigl( (\hypspace, \mathbf{h}) \bigr)$.
\item 
Minimax MTL as \emph{$\ell_\infty$ MTL}: 

\qquad $\displaystyle
\qquad \inf_{\hypspace \in \mathbb{H}} \inf_{\mathbf{h} \in \hypspace^T} \max_{t \in [T]} \hat{\ell}(h_t) + \Omega \bigl( (\hypspace, \mathbf{h}) \bigr) 
\qquad \,\,\,\,\, \equiv \,\,
\inf_{\hypspace \in \mathbb{H}} \inf_{\mathbf{h} \in \hypspace^T} \| \hat{\bm{\ell}}(\mathbf{h}) \|_\infty + \Omega \bigl( (\hypspace, \mathbf{h}) \bigr)$.
\item 
A new formulation, \emph{$\ell_2$ MTL}: 

\qquad $\displaystyle
\inf_{\hypspace \in \mathbb{H}} \inf_{\mathbf{h} \in \hypspace^T} 
\Bigl(\frac{1}{T} \sum_{t \in [T]} \bigl(\hat{\ell}(h_t)\bigr)^2 \Bigr)^{1/2}
 + \Omega \bigl( (\hypspace, \mathbf{h}) \bigr) \,\,
\equiv \,\,
\inf_{\hypspace \in \mathbb{H}} \inf_{\mathbf{h} \in \hypspace^T} \frac{1}{\sqrt{T}} \| \hat{\bm{\ell}}(\mathbf{h}) \|_2 + \Omega \bigl( (\hypspace, \mathbf{h}) \bigr)$.
\end{list}

A natural question is why one might consider minimizing $\ell_p$-norms of the empirical risks vector for $1 < p < \infty$, as in $\ell_2$ MTL. 
The contour of the $\ell_1$-norm of the empirical risks evenly trades off empirical risks between different tasks; however, it has been observed that overfitting often happens near the end of learning, rather than the beginning \cite{leroux2008topmoumoute}. 
More precisely, when the empirical risk is high, the gradient of the empirical risk (taken with respect to the parameter $(\hypspace, \mathbf{h})$) is likely to have positive inner product with the gradient of the true risk. Therefore, given a candidate solution with a corresponding vector of empirical risks, a sensible strategy is to take a step in solution space which places more emphasis on tasks with higher empirical risk. This strategy is particularly appropriate when the class of learners has high capacity relative to the amount of available data. This observation sets the foundation for an approach that minimizes norms of the empirical risks.

In this work, we also discuss an interesting subset of the loss-compositional paradigm which does not fit into $\ell_p$ MTL; this subfamily embodies a continuum of relaxations of minimax MTL.

\paragraph{$\bm{\alpha}$-minimax MTL.}
In some cases, minimizing the maximum loss can exhibit certain disadvantages because the maximum loss is not robust to situations when a small fraction of the tasks are fundamentally harder than the remaining tasks. Consider the case when the empirical risk for each task in this small fraction can not be reduced below a level $u$. Rather than rigidly minimizing the maximum loss, a more robust alternative is to minimize the maximize loss in a soft way. Intuitively, the idea is to ensure that most tasks have low empirical risk, but a small fraction of tasks are permitted to have higher loss. We formalize this as \emph{$\alpha$-minimax MTL}, via the relaxed objective:
\begin{align}
\begin{aligned}
&\underset{\hypspace \in \mathbb{H}, \mathbf{h} \in \hypspace^T}{\text{minimize}}
&& \min_{b \geq 0} \Bigl\{ b + \frac{1}{\alpha} \sum_{t \in [T]} \max\{0, \hat{\ell}_t(h_t) - b\} \Bigr\} + \Omega \bigl( (\hypspace, \mathbf{h}) \bigr) .
\end{aligned}
\end{align}
In the above, $\phi$ from the loss-compositional paradigm \eqref{eqn:paradigm} is a variational function of the empirical risks vector. 
The above optimization problem is equivalent to the perhaps more intuitive problem:
\begin{align}
   \underset{\hypspace \in \mathbb{H}, \mathbf{h} \in \hypspace^T,b \geq 0,\bm{\xi} \geq 0}{\text{minimize}}  \quad b + \frac{1}{\alpha} \sum_{t \in [T]} \xi_t + \Omega \bigl( (\hypspace, \mathbf{h}) \bigr) \qquad
   \text{subject to} \quad
    \hat{\ell}_t(h_t) \leq  b + \xi_t, \; t \in [T]. 
\end{align}
Here, $b$ plays the role of the relaxed maximum, and each $\xi_t$'s deviation from zero indicates the deviation from the (loosely enforced) maximum. We expect $\bm{\xi}$ to be sparse. 

To help understand how $\alpha$ affects the learning problem, let us consider a few cases:
\begin{enumerate}[(1)]
\item 
When $\alpha > T$, the optimal value of $b$ is zero, and the problem is equivalent to classical MTL. To see this, note that for a given candidate solution with $b > 0$ the objective always can be reduced by reducing $b$ by some $\varepsilon$ and increasing each $\xi_t$ by the same $\varepsilon$. 
\item 
Suppose one task is much harder than all the other tasks (e.g. an outlier task), and its empirical risk is separated from the maximum empirical risk of the other tasks by $\rho$. Let $1 < \alpha < 2$; now, at the optimal hard maximum solution (where $\bm{\xi} = \mathbf{0}$), the objective can be reduced by increasing one of the $\xi_t$'s by $\rho$ and decreasing $b$ by $\rho$. Thus, the objective can focus on minimizing the maximum risk of the set of $T - 1$ easier tasks. 
In this special setting, this argument can be extended to the more general case $k < \alpha < k+1$ and $k$ outlier tasks, for $k \in [T]$. 
\item 
As $\alpha$ approaches $0$, we recover the hard maximum case of minimax MTL.
\end{enumerate}

This work focuses on $\alpha$-minimax MTL with 
$\alpha = 2 / (\lceil0.1 T + 0.5 \rceil^{-1} + \lceil 0.1 T + 1.5 \rceil^{-1})$
i.e. the harmonic mean of 
$\lceil 0.1 T + 0.5 \rceil$ and $\lceil 0.1 T + 1.5 \rceil$.
The reason for this choice is that in the idealized case (2) above, for large $T$ this setting of $\alpha$ makes the relaxed maximum consider all but the hardest 10\% of the tasks. We also try the 20\% level (i.e. $0.2 T$ replacing $0.1 T$ in the above).

\paragraph{Models.}
We now provide examples of how specific models fit into this framework. 
We consider two convex multi-task learning formulations: Evgeniou and Pontil's regularized multi-task learning (the \emph{EP model}) \cite{evgeniou2004regularized} and Argyriou, Evgeniou, and Pontil's convex multi-task feature learning (the \emph{AEP model}) \cite{argyriou2008convex}. 
The EP model is a linear model with a shared parameter $v_0 \in \real^d$ and task-specific parameters $v_t \in \real^d$ (for $t \in [T]$). Evgeniou and Pontil presented this model as
\begin{align}
\textstyle
\min_{v_0, \{v_t\}_{t \in [T]}} \sum_{t \in [T]} \sum_{i=1}^m \ell(y_{t,i}, \langle v_0 + v_t, x_{t,i} \rangle) + \lambda_0 \|v_0\|^2 + \frac{\lambda_1}{T} \sum_{t \in [T]} \|v_t\|^2 ,
\end{align}
for $\ell$ the hinge loss or squared loss. 
This can be set in the new paradigm via 
$\mathbb{H} = \{\hypspace_{v_0} \mid v_0 \in \real^d\}$, $\hypspace_{v_0} = \{h: x \mapsto \langle v_0 + v_t, x \rangle \mid v_t \in \real^d\}$, and $\hat{\ell}_t(h_t) = \frac{1}{m} \sum_{i=1}^m \ell \bigl( y_{t,i}, \langle v_0 + v_t, x_{t,i} \rangle \bigr)$.

The AEP model minimizes the task-wise average loss with the trace norm (nuclear norm) penalty:
\begin{align}
\textstyle
\min_W \sum_t \sum_{i=1}^m \ell(y_{t,i}, \langle W_t, x_{t,i} \rangle) + \lambda \|W\|_{\mathrm{tr}} ,
\end{align}
where $\|\cdot\|_{\mathrm{tr}}: W \mapsto \sum_i \sigma_i (W)$ is the trace norm.  
In the new paradigm, 
$\mathbb{H}$ is a set where each element is a $k$-dimensional subspace of linear estimators (for $k \ll d$). Each $h_t = W_t$ in some $\hypspace \in \mathbb{H}$ lives in $\hypspace$'s corresponding low-dimensional subspace. Also, $\hat{\ell}_t(h_t) = \frac{1}{m} \sum_{i=1}^m \ell \bigl( y_{t,i}, \langle h_t, x_{t,i} \rangle \bigr)$. 

For easy empirical comparison between the various MTL formulations from the paradigm, at times it will be convenient to use constrained formulations of the EP and AEP model. If the regularized forms are used, a fair comparison of the methods warrants plotting results according to the size of the optimal parameter found (i.e. $\|W\|_{\mathrm{tr}}$ for AEP). For EP, the constrained form is:
\begin{align*}
\textstyle
\min_{v_0, \{v_t\}_{t \in [T]}} \sum_{t \in [T]} \sum_{i=1}^m \ell(y_{t,i}, \langle v_0 + v_t, x_{t,i} \rangle) 
\quad \text{subject to} \,\, \|v_0\| \leq \tau_0, \,\, \|v_t\| \leq \tau_1 \, \text{for } t \in [T] .
\end{align*}
For AEP, the constrained form is: \quad
$\min_W \sum_t \sum_{i=1}^m \ell(y_{t,i}, \langle W_t, x_{t,i} \rangle)
\quad \text{subject to} \,\, \|W\|_{\mathrm{tr}} \leq r $.

\section{Empirical evaluation}
\label{sec:experiments}
We consider four learning problems; the first three involve regression 
(MTL model in parentheses):
\vspace{-2mm}
\begin{list}{\labelitemi}{\leftmargin=2em \parsep=0pt \itemsep=0.5pt}
\item A synthetic dataset composed from \emph{two modes} of tasks (EP model),
\item The \emph{school} dataset from the Inner London Education Authority (EP model),
\item The conjoint analysis \emph{personal computer} ratings dataset \footnote{This data, collected at the University of Michigan MBA program, generously was provided by Peter Lenk.} \cite{lenk1996hierarchical} (AEP model).
\end{list}
\vspace{-2mm}
The fourth problem is multi-class classification from the \emph{MNIST} digits dataset \cite{lecun1998gradient} with a reduction to multi-task learning using a tournament of pairwise (binary) classifiers. We use the AEP model. 
Given data, each problem involved a choice of MTL formulation (e.g. minimax MTL), model (EP or AEP), and choice of regularized versus constrained. All the problems were solved using just a few lines of code using \texttt{CVX} \cite{grant2011cvx,grant2008graph}. 
In this work, we considered convex multi-task learning formulations in order to make clear statements about the optimal solutions attained for various learning problems.

\begin{figure}[t]
\centering
\parbox{43mm}{
\includegraphics[width=43mm]{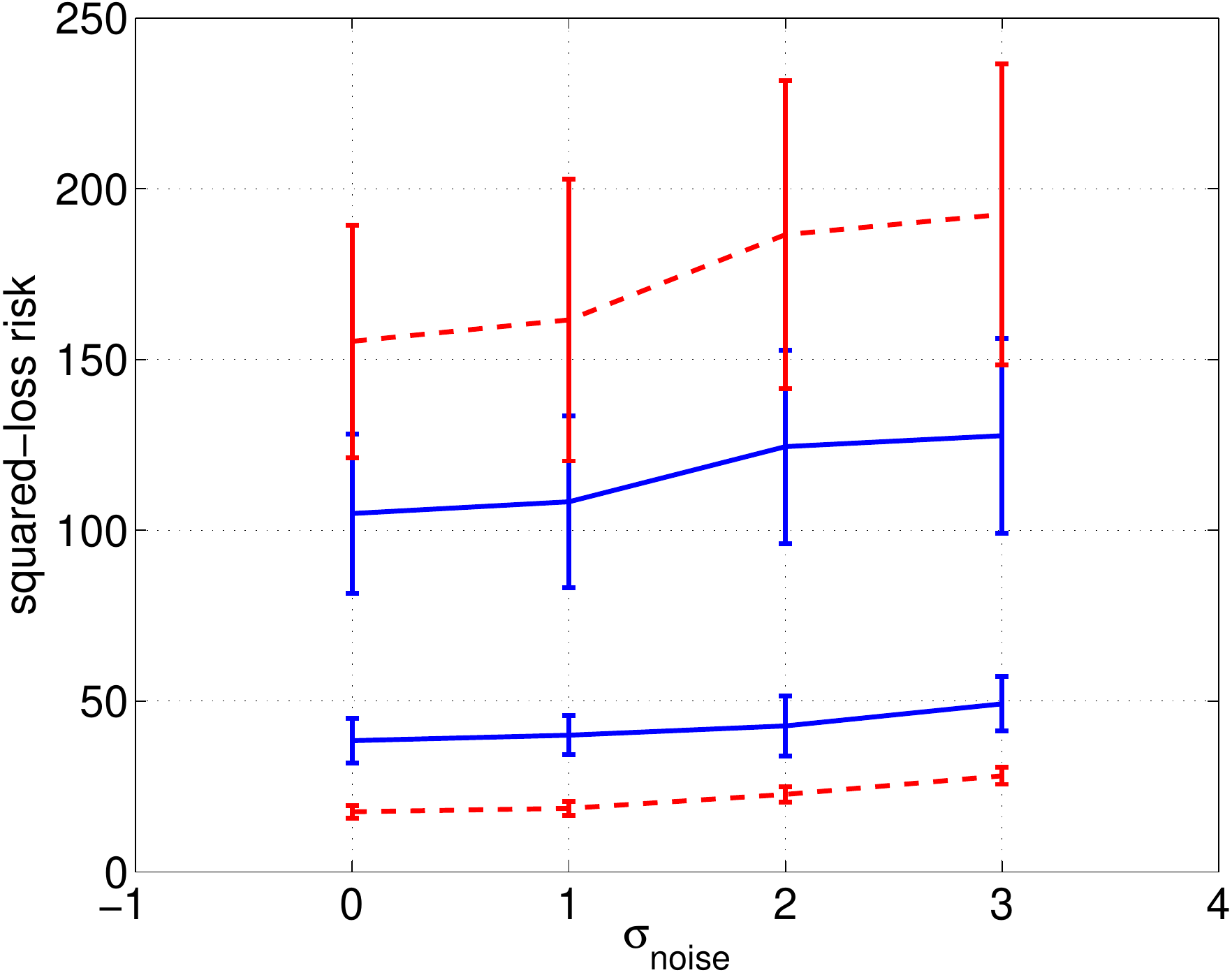}
}
\quad
\begin{minipage}{43mm}
\includegraphics[width=43mm]{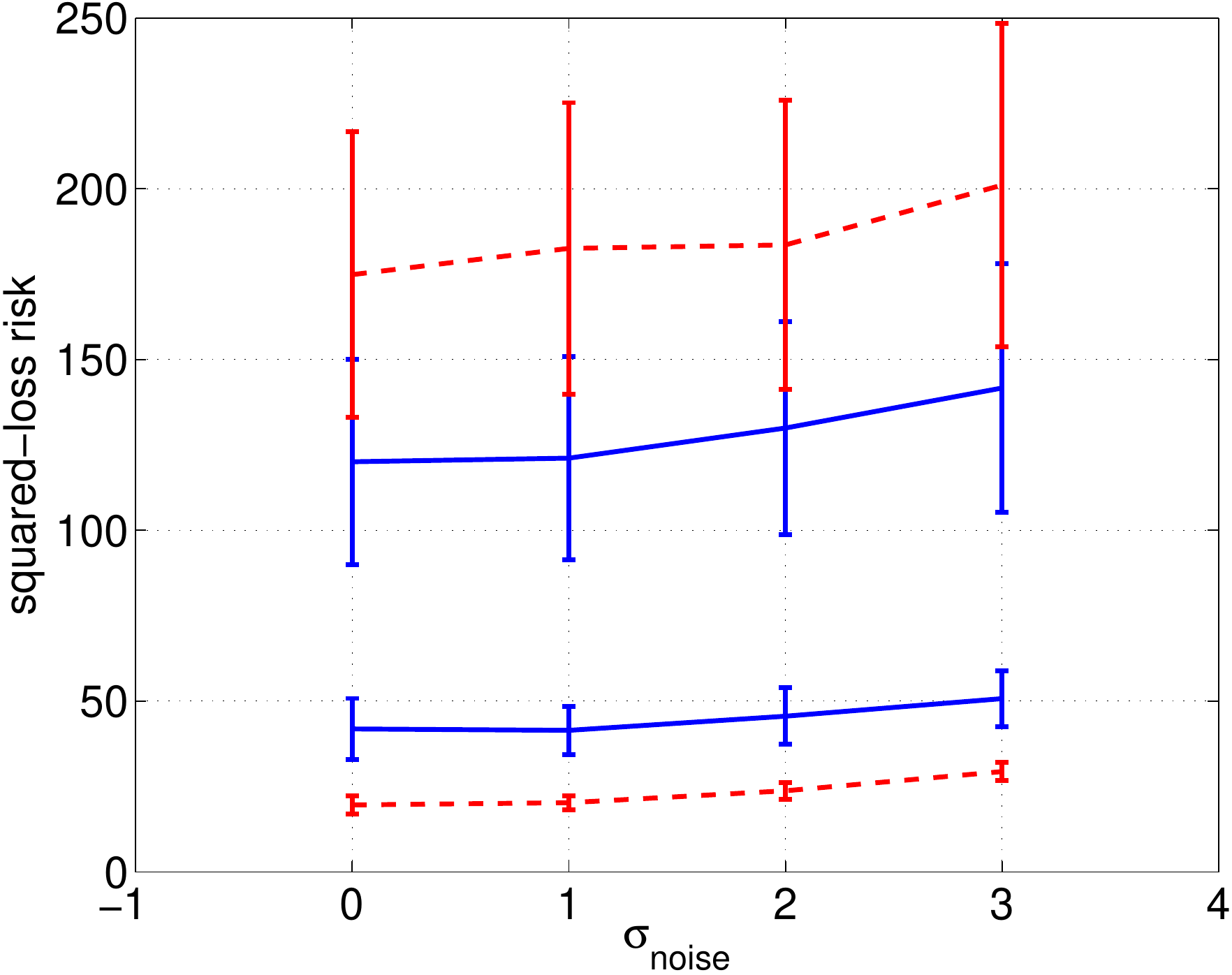}
\end{minipage}
\quad
\begin{minipage}{43mm}
\includegraphics[width=43mm]{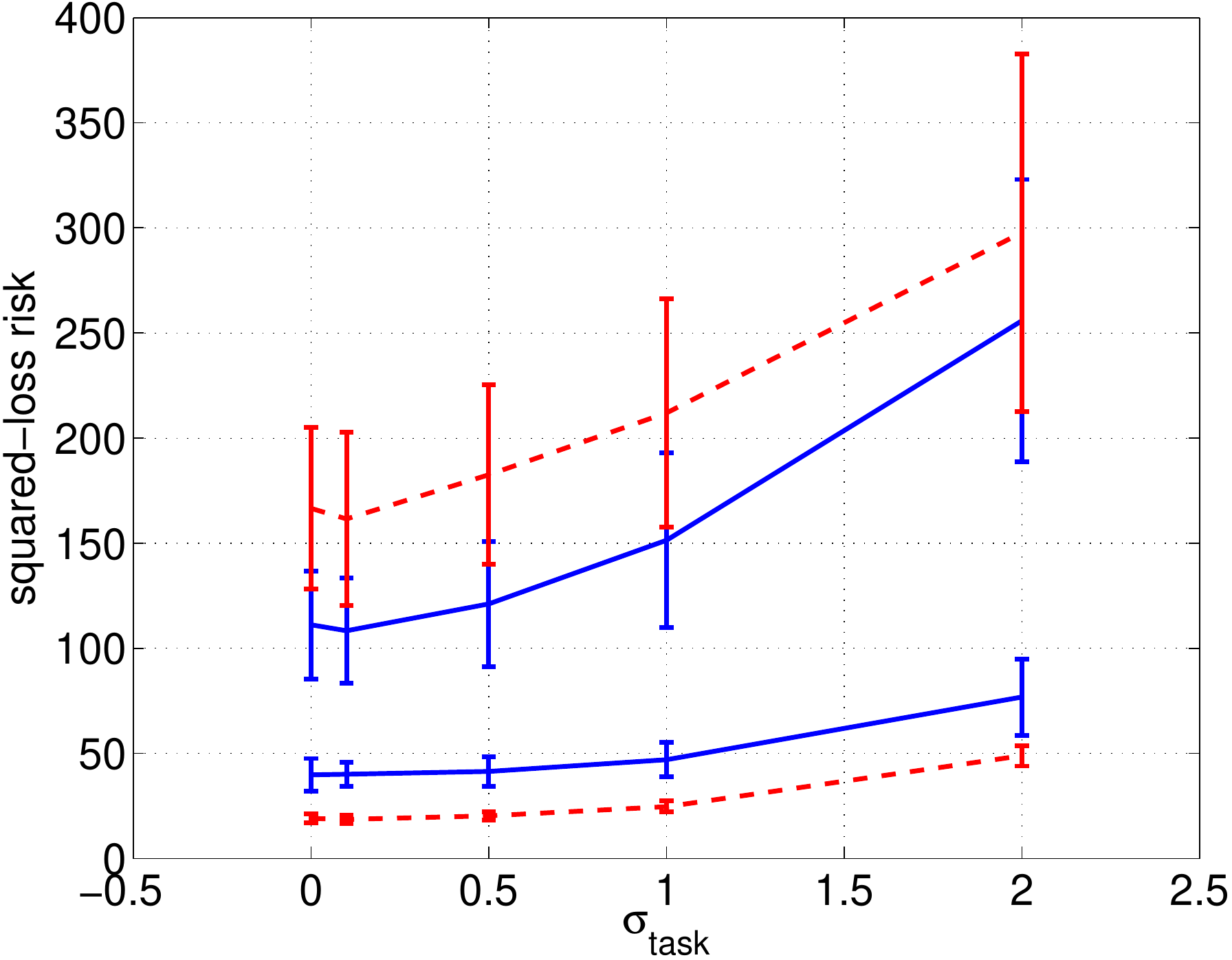}
\end{minipage}
\vspace{-3mm}
\caption{\small
Max $\ell_2$-risk (Top two lines) and mean $\ell_2$-risk (Bottom two lines). 
At Left and Center: $\ell_2$-risk vs noise level, for $\sigma_{\mathrm{task}} = 0.1$ and $\sigma_{\mathrm{task}} = 0.5$ respectively. At Right: $\ell_2$-risk vs task variation, for $\sigma_{\mathrm{noise}} = 0.1$. Dashed red is $\ell_1$, dashed blue is minimax. Error bars indicate one standard deviation. MTL results (not shown) were similar to LTL results (shown), with MTL-LTL relative difference  below 6.8\%  for all points plotted.
}
\label{fig:two-modes-squared-loss}
\end{figure}

\paragraph{Two modes.}
The two modes regression problem consists of 50 linear prediction tasks for the first type of task and 5 linear prediction tasks for the second task type. The true parameter for the first task type is a vector $\mu$ drawn uniformly from the sphere of radius 5; the true parameter for the second task type is $-2 \mu$. Each task is drawn from an isotropic Gaussian with mean taken from the task type and the standard deviation of all dimensions set to $\sigma_{\mathrm{task}}$. Each data point for each task is drawn from a product of 10 standard normals (so {\small $x_{t,i} \in \real^{10}$}). The targets are generated according to {\small $\langle W_t, x_{t,i} \rangle + \varepsilon_t$}, where the $\varepsilon_t$'s are iid univariate centered normals with standard deviation $\sigma_{\mathrm{noise}}$. 
We fixed $\tau_0$ to a large value (in this case, $\tau_0 = 10$ is sufficient since the mean for the largest task fits into a ball of radius 10) and $\tau_1$ to a small value ($\tau_1 = 2$). 
We compute the average mean and maximum test error over 100 instances of the 55-task multi-task problem. Each task's training set and test set are 5 and 15 points respectively. The average maximum (mean) test error is the 100-experiment-average of the task-wise maximum (mean) of the $\ell_2$ risks. For each LTL experiment, 55 new test tasks were drawn using the same $\mu$ as from the training tasks. 

Figure \ref{fig:two-modes-squared-loss} shows a tradeoff: when each task group is fairly homogeneous (left and center plots), minimax is better at minimizing the maximum of the test risks while $\ell_1$ is better at minimizing the mean of the test risks. 
As task homogeneity decreases (right plot), the gap in performance closes with respect to the maximum of the test risks and remains roughly the same with respect to the mean.

\begin{figure}[h]
\centering
\parbox{68mm}{
\includegraphics[width=68mm]{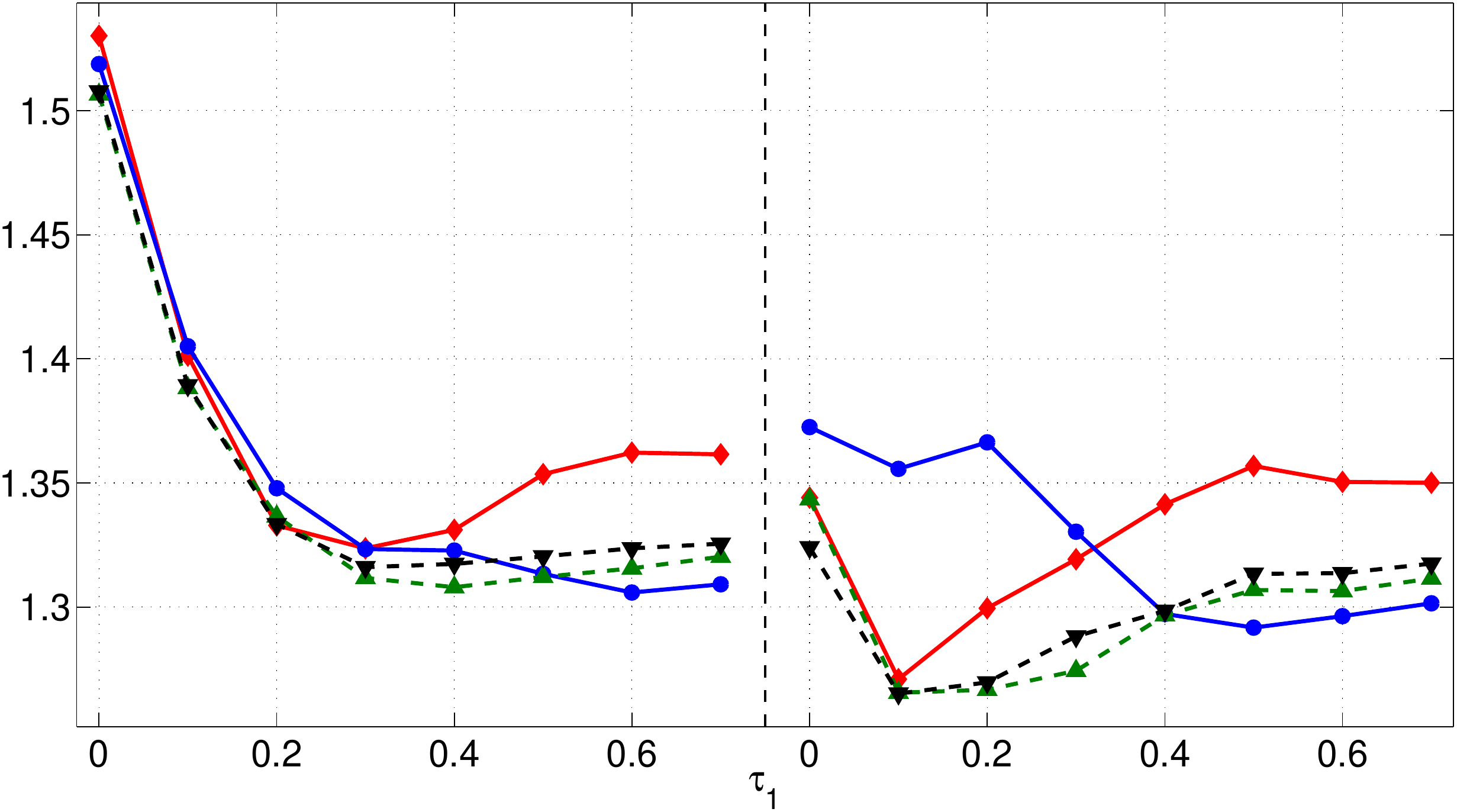}
}
\,\,
\begin{minipage}{68mm}
\includegraphics[width=68mm]{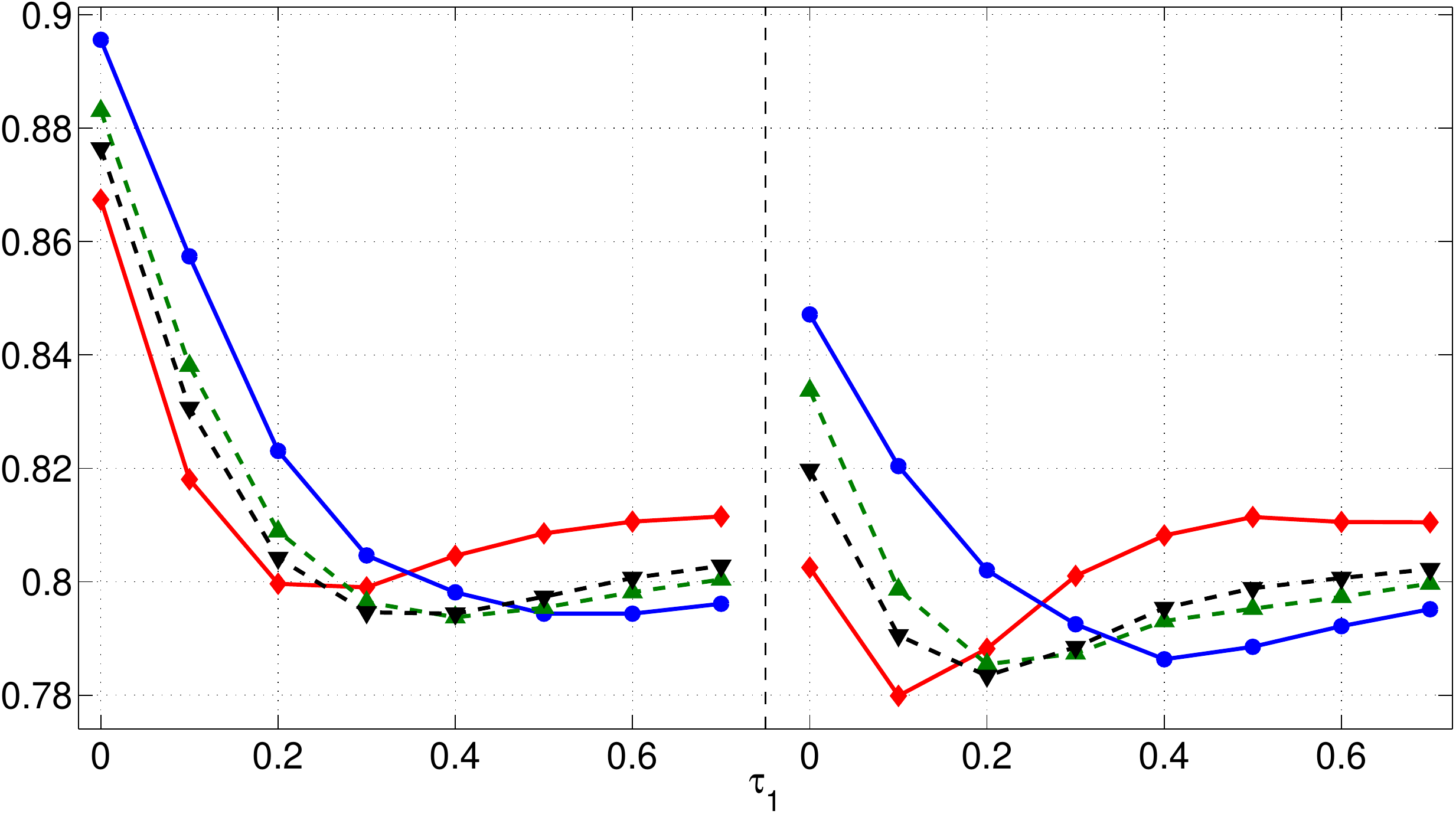}
\end{minipage}
\vspace{-3mm}
\caption{\small
Maximum RMSE (Left) and normalized mean RMSE (Right) versus task-specific parameter bound $\tau_1$,  for shared parameter bound $\tau_0$ fixed. In each figure, Left section is $\tau_0$ is 0.2 and Right section is $\tau_0 = 0.6$. 
Solid red {\scriptsize \textcolor{red}{$\blacklozenge$}} is $\ell_1$, solid blue \textcolor{blue}{$\bullet$} is minimax, dashed green \textcolor{OliveGreen}{$\blacktriangle$} is $(0.1 T)$-minimax, dashed black $\blacktriangledown$ is $(0.2 T)$-minimax. The results for $\ell_2$ MTL were visually identical to $\ell_1$ MTL and hence were not plotted. 
}
\label{fig:school-RMSE}
\end{figure}

\vspace{-5mm}

\paragraph{School.}
The school dataset has appeared in many previous works \cite{goldstein1991multilevel,bakker2003task,evgeniou2007convex}. For brevity we just say the goal is to predict student test scores using certain student-level features. 
Each school is treated as a separate task. 
We report both the task-wise maximum of the root mean square error (RMSE) and the taskwise-mean of the RMSE (normalized by number of points per task, as in previous works).

The results (see Figure \ref{fig:school-RMSE}) demonstrate that when the learner has moderate shared capacity $\tau_0$ and high task-specific capacity $\tau_1$, minimax MTL outperforms $\ell_1$ MTL for the max objective; additionally, for the max objective in almost all parameter settings $(0.1 T)$-minimax and $(0.2 T)$-minimax MTL outperform $\ell_1$ MTL, and they also outperform minimax MTL when the task-specific capacity $\tau_1$ is not too large. We hypothesize that minimax MTL performs the best in the high$-\tau_1$ regime because stopping learning once the maximum of the empirical risks cannot be improved invokes early stopping and its built-in regularization properties (see e.g. \cite{murata1999statistical}). 
Interestingly, for the normalized mean RMSE objective, both minimax relaxations are competitive with $\ell_1$ MTL; however, when the shared capacity $\tau_0$ is high (right section, right plot), $\ell_1$ MTL performs the best. For high task-specific capacity $\tau_1$, minimax MTL and its relaxations again seem to resist overfitting compared to $\ell_1$ MTL.

\begin{wrapfigure}{r}{83mm}
  \vspace{-15pt}
  \begin{center}
    \parbox{41mm}{
      \includegraphics[width=41mm]{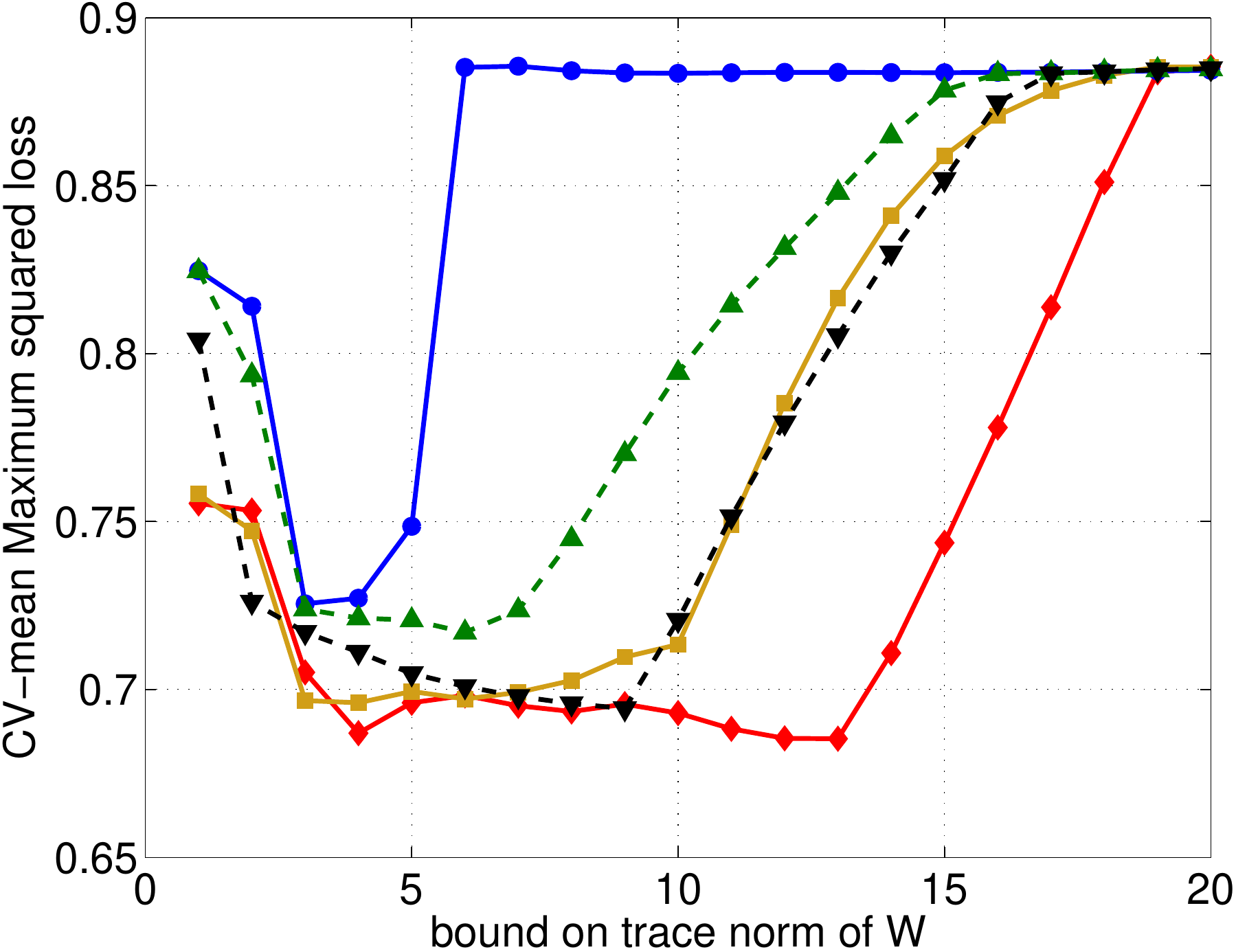}
    }
    \begin{minipage}{41mm}
      \includegraphics[width=41mm]{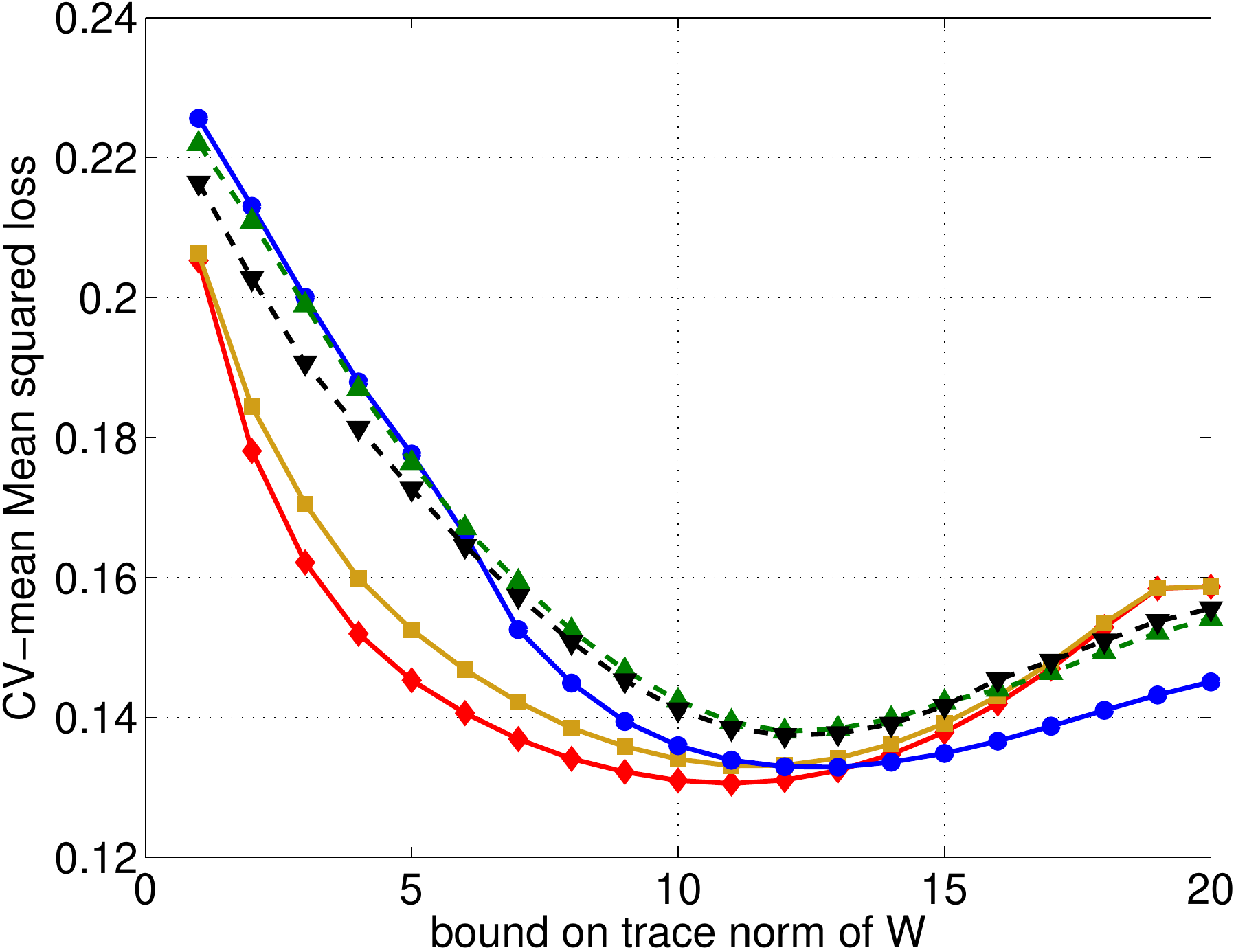}
    \end{minipage}
  \end{center}
  \begin{center}
    \parbox{41mm}{    \includegraphics[width=41mm]{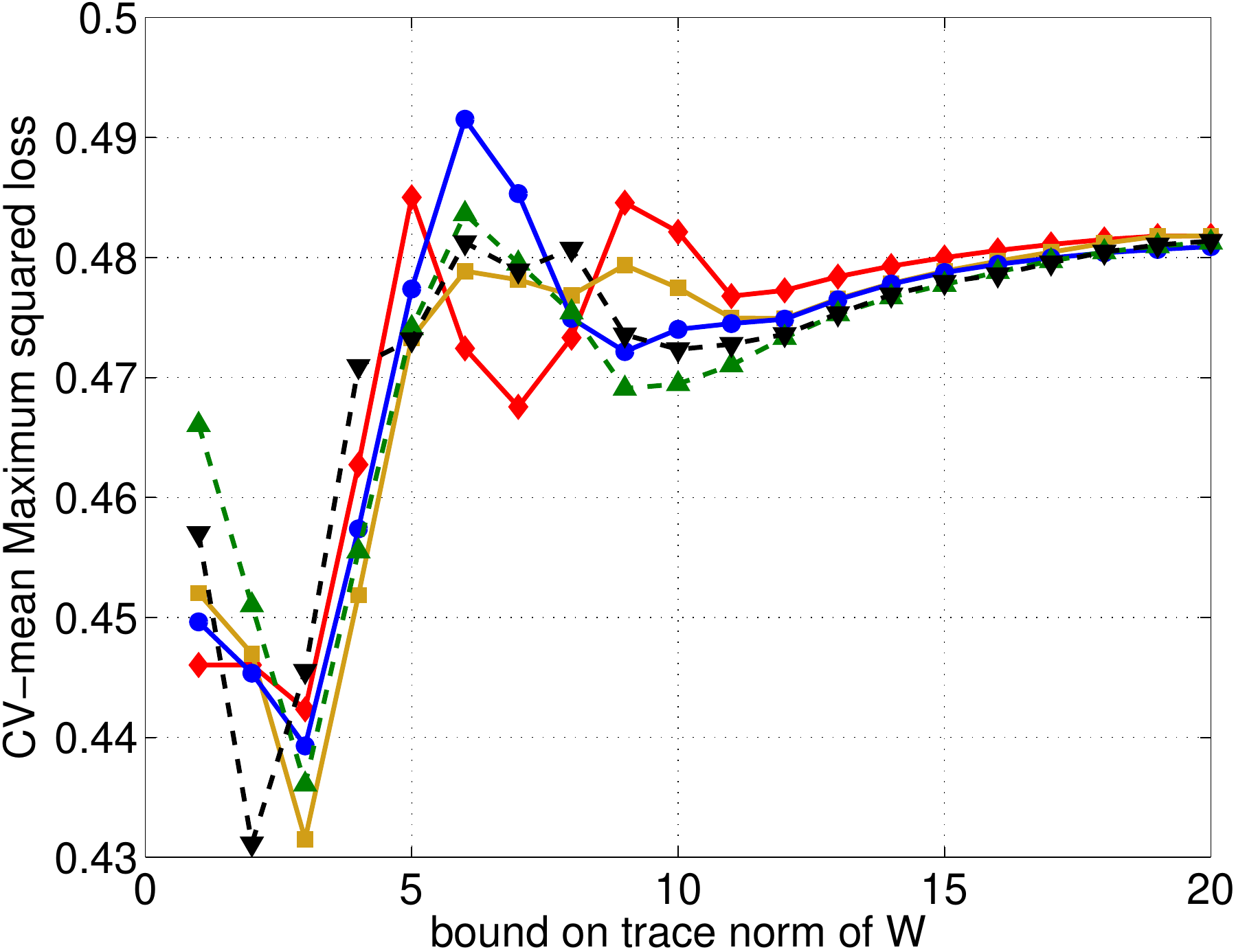}
    }
    \begin{minipage}{41mm}  \includegraphics[width=41mm]{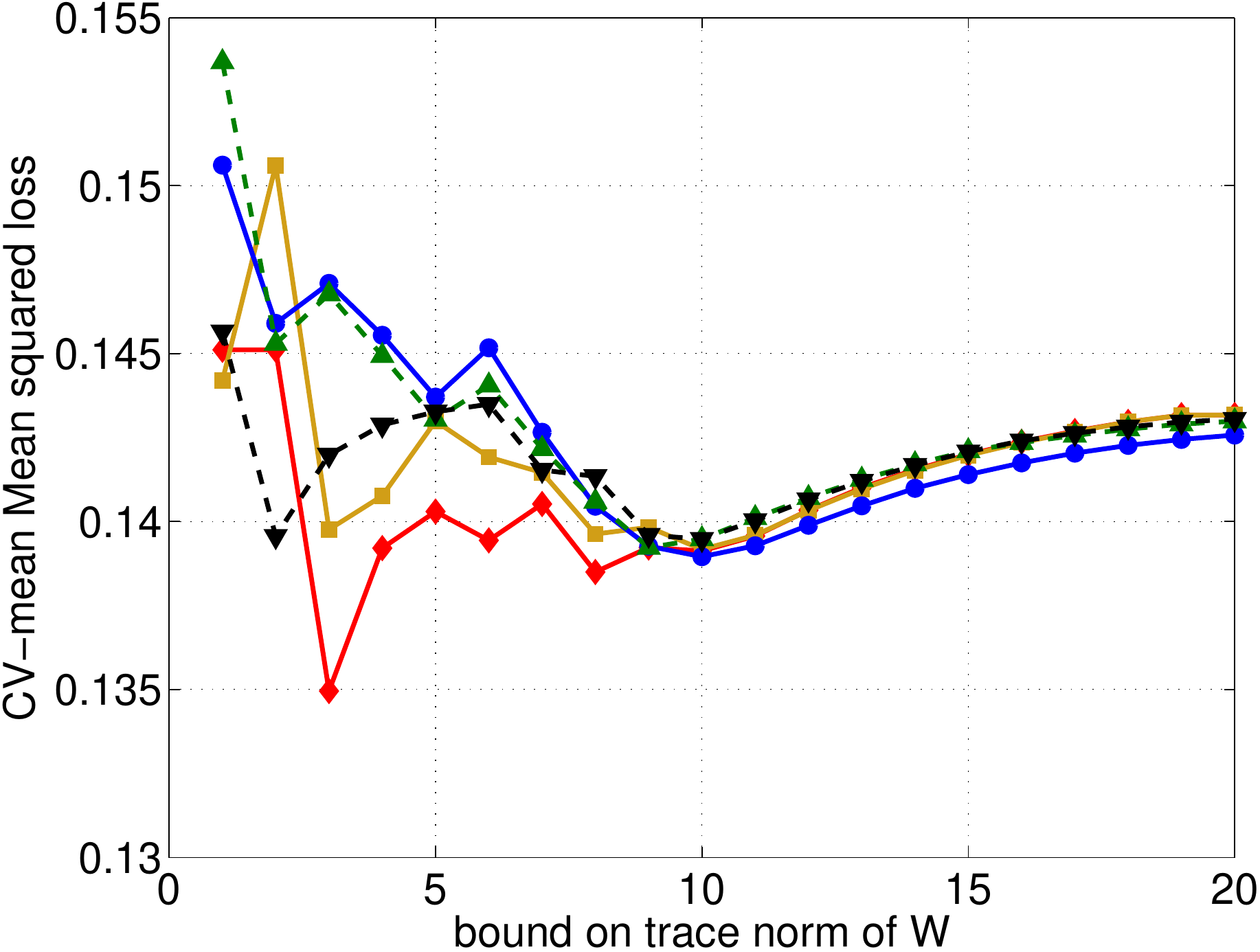}
    \end{minipage}
  \end{center}
\vspace{-3mm}
  \caption{\small 
MTL (Top) and LTL (Bottom). Maximum $\ell_2$ risk (Left) and Mean $\ell_2$ risk (Right) vs bound on $\|W\|_{\mathrm{tr}}$. LTL used 10-fold cross-validation (10\% of tasks left out in each fold). 
Solid red {\scriptsize \textcolor{red}{$\blacklozenge$}} is $\ell_1$, solid blue \textcolor{blue}{$\bullet$} is minimax, dashed green \textcolor{OliveGreen}{$\blacktriangle$} is $(0.1 T)$-minimax, dashed black $\blacktriangledown$ is $(0.2 T)$-minimax, solid gold {\scriptsize \textcolor{BurntOrange}{$\blacksquare$}} is $\ell_2$. 
}
\label{fig:computer}
\vspace{6mm}
\end{wrapfigure}

\vspace{-1mm}

\paragraph{Personal computer.}
The personal computer dataset is composed of 189 human subjects 
each of which rated on a 0-10 scale the same 20 computers (16 training, 4 test). Each computer has 13 binary features (amount of memory, screen size, price, etc.). 

The results are shown in Figure \ref{fig:computer}. In the MTL setting, for both the maximum RMSE objective and the mean RMSE objective, $\ell_1$ MTL appears to perform the best. When the trace norm of $W$ is high, minimax MTL displays resistance to overfitting and obtains the lowest mean RMSE. In the LTL setting for the maximum RMSE objective, $\ell_2$, minimax, and $(0.1 T)$-minimax MTL all outperform $\ell_1$ MTL. For the mean RMSE, $\ell_1$ MTL obtains the lowest risk for almost all parameter setttings.

\begin{wrapfigure}{r}{55mm}
  \vspace{-17pt}
  \begin{center}
    \includegraphics[width=55mm]{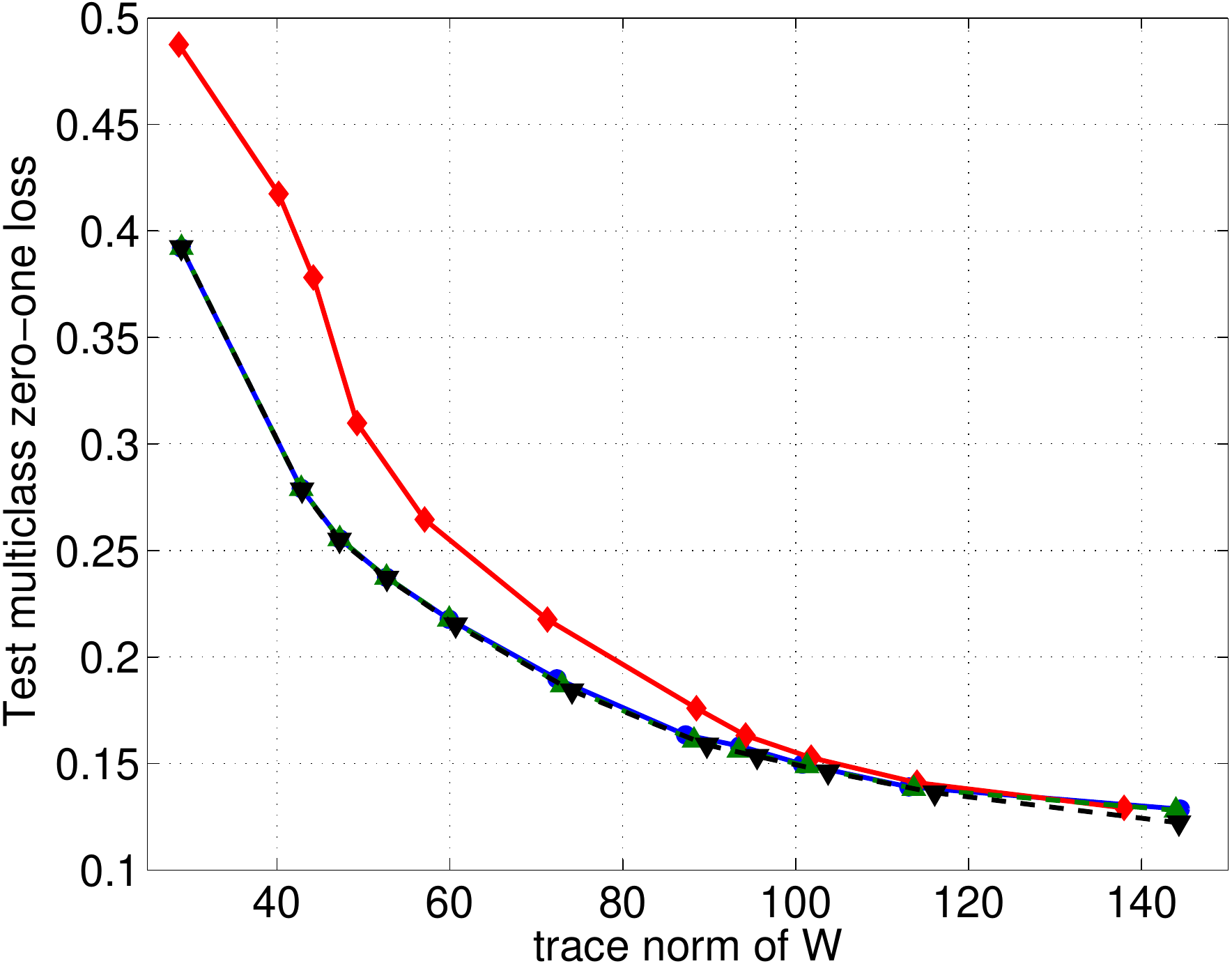}
  \end{center}
  \vspace{-15pt}
  \caption{\small
Test multiclass 0-1 loss vs $\|W\|_{\mathrm{tr}}$. Solid red is $\ell_1$ MTL, solid blue is minimax, dashed green is $(0.1 T)$-minimax, dashed black is $(0.2 T)$-minimax. Regularized AEP used for speed and trace norm of $W$'s computed, so samples differ per curve.
}
  \label{fig:mnist-test}
  \vspace{-25pt}
\end{wrapfigure}

\paragraph{MNIST.}
The MNIST task is a 10-class problem; we approach it via a reduction to a tournament of 45 binary classifiers trained via the AEP model. The dimensionality was reduced to 50 using principal component analysis (computed on the full training set), and only the first 2\% of each class's training points was used for training. 

Intuitively, the performance of the tournament tree of binary classifiers can only be as accurate as its paths, and the accuracy of each path depends on the accuracy of the nodes. Hence, our hypothesis is that minimax MTL should outperform $\ell_1$ MTL. The results in Figure \ref{fig:mnist-test} confirm our hypothesis. Minimax MTL outperforms $\ell_1$ MTL when the capacity $\|W\|_{\mathrm{tr}}$ is somewhat limited, with the gap widening as the capacity decreases. 
Furthermore, at every capacity minimax MTL is competitive with $\ell_1$ MTL.

\section{Discussion}
We have established a continuum of formulations for MTL which recovers as special cases classical MTL and the newly formulated minimax MTL. In between these extreme points lies a continuum of relaxed minimax MTL formulations. More generally, we introduced a loss-compositional paradigm that operates on the vector of empirical risks, inducing the additional $\ell_p$ MTL paradigms. The empirical evaluations indicate that $\alpha$-minimax MTL at either the 10\% or 20\% level often outperform $\ell_1$ MTL in terms of the maximum test risk objective and sometimes even in the mean test risk objective. 
All the minimax or $\alpha$-minimax MTL formulations exhibit a built-in safeguard against overfitting in the case of learning with a model that is very complex relative to the available data. 

Although efficient algorithms may make the various new MTL learning formulations practical for large problems, a proper effort to develop fast algorithms in this setting would have detracted from the main point of this first study. A good direction for the future is to obtain efficient algorithms for minimax and $\alpha$-minimax MTL. In fact, such algorithms might have applications beyond MTL and even machine learning. Another area ripe for exploration is to establish more general learning bounds for minimax MTL and to extend these bounds to $\alpha$-minimax MTL.

\newpage
{
\bibliographystyle{plain}
\bibliography{minimax_mtl}

\begin{thebibliography}{10}

\bibitem{argyriou2008convex}
A.~Argyriou, T.~Evgeniou, and M.~Pontil.
\newblock Convex multi-task feature learning.
\newblock {\em Machine Learning}, 73(3):243--272, 2008.

\bibitem{bakker2003task}
B.~Bakker and T.~Heskes.
\newblock Task clustering and gating for bayesian multitask learning.
\newblock {\em The Journal of Machine Learning Research}, 4:83--99, 2003.

\bibitem{bartlett2002rademacher}
Peter~L. Bartlett and Shahar Mendelson.
\newblock {R}ademacher and {G}aussian complexities: Risk bounds and structural
  results.
\newblock {\em Journal of Machine Learning Research}, 3:463--482, 2002.

\bibitem{baxter2000model}
J.~Baxter.
\newblock A model of inductive bias learning.
\newblock {\em Journal of Artificial Intelligence Research}, 12(1):149--198,
  2000.

\bibitem{evgeniou2004regularized}
T.~Evgeniou and M.~Pontil.
\newblock Regularized multi--task learning.
\newblock In {\em Proceedings of the tenth ACM SIGKDD international conference
  on Knowledge discovery and data mining}, pages 109--117. ACM, 2004.

\bibitem{evgeniou2007convex}
T.~Evgeniou, M.~Pontil, and O.~Toubia.
\newblock A convex optimization approach to modeling consumer heterogeneity in
  conjoint estimation.
\newblock {\em Marketing Science}, 26(6):805--818, 2007.

\bibitem{goldstein1991multilevel}
H.~Goldstein.
\newblock Multilevel modelling of survey data.
\newblock {\em Journal of the Royal Statistical Society. Series D (The
  Statistician)}, 40(2):235--244, 1991.

\bibitem{grant2008graph}
M.~Grant and S.~Boyd.
\newblock Graph implementations for nonsmooth convex programs.
\newblock In V.~Blondel, S.~Boyd, and H.~Kimura, editors, {\em Recent Advances
  in Learning and Control}, Lecture Notes in Control and Information Sciences,
  pages 95--110. Springer-Verlag Limited, 2008.

\bibitem{grant2011cvx}
M.~Grant and S.~Boyd.
\newblock {CVX}: Matlab software for disciplined convex programming, version
  1.21, April 2011.

\bibitem{lecun1998gradient}
Y.~LeCun, L.~Bottou, Y.~Bengio, and P.~Haffner.
\newblock Gradient-based learning applied to document recognition.
\newblock {\em Proceedings of the IEEE}, 86(11):2278--2324, 1998.

\bibitem{lenk1996hierarchical}
P.J. Lenk, W.S. DeSarbo, P.E. Green, and M.R. Young.
\newblock Hierarchical bayes conjoint analysis: Recovery of partworth
  heterogeneity from reduced experimental designs.
\newblock {\em Marketing Science}, pages 173--191, 1996.

\bibitem{maurer2009transfer}
A.~Maurer.
\newblock Transfer bounds for linear feature learning.
\newblock {\em Machine learning}, 75(3):327--350, 2009.

\bibitem{murata1999statistical}
N.~Murata and S.~Amari.
\newblock Statistical analysis of learning dynamics.
\newblock {\em Signal Processing}, 74(1):3--28, 1999.

\bibitem{leroux2008topmoumoute}
Nicolas~Le Roux, Pierre-Antoine Manzagol, and Yoshua Bengio.
\newblock Topmoumoute online natural gradient algorithm.
\newblock In J.C. Platt, D.~Koller, Y.~Singer, and S.~Roweis, editors, {\em
  Advances in Neural Information Processing Systems 20}, pages 849--856. MIT
  Press, Cambridge, MA, 2008.

\bibitem{yu2009large}
K.~Yu, J.~Lafferty, S.~Zhu, and Y.~Gong.
\newblock Large-scale collaborative prediction using a nonparametric random
  effects model.
\newblock In {\em Proceedings of the 26th Annual International Conference on
  Machine Learning}, pages 1185--1192. ACM, 2009.

\bibitem{zhang2011generalizing}
L.~Zhang, D.~Agarwal, and B.C. Chen.
\newblock Generalizing matrix factorization through flexible regression priors.
\newblock In {\em Proceedings of the fifth ACM conference on Recommender
  systems}, pages 13--20. ACM, 2011.

\end{thebibliography}
}

\end{document}